\documentclass[conference]{IEEEtran}
\IEEEoverridecommandlockouts
\usepackage{cite}
\usepackage{amsmath,amssymb,amsfonts}
\usepackage{algorithmic}
\usepackage{graphicx}
\usepackage{textcomp}
\usepackage{xcolor}
\usepackage{marvosym}

\usepackage{amsmath,amsfonts}
\usepackage{algorithmic}
\usepackage{algorithm}
\usepackage{array}
\usepackage[caption=false,font=normalsize,labelfont=sf,textfont=sf]{subfig}
\usepackage{textcomp}
\usepackage{stfloats}
\usepackage{url}
\usepackage{verbatim}
\usepackage{graphicx}
\usepackage{cite}
\usepackage{cleveref}
\crefname{figure}{Fig.}{Figs.}
\crefname{table}{Tab.}{Tabs.}
\crefname{section}{Sec.}{Secs.}
\crefname{equation}{Eq.}{Eqs.}  
\Crefname{figure}{Fig.}{Figs.}
\Crefname{table}{Tab.}{Tabs.}
\Crefname{section}{Sec.}{Secs.}
\Crefname{equation}{Eq.}{Eqs.} 

\newcommand{\methodName}{TUGS}
\usepackage{multirow}
\usepackage{xcolor}
\usepackage{colortbl}
\usepackage{bm}
\definecolor{lightred}{RGB}{255,200,200}
\definecolor{lightorange}{RGB}{255,215,200}
\definecolor{lightyellow}{RGB}{255,255,200}
\definecolor{skyblue}{rgb}{0.53, 0.81, 0.98}
\definecolor{lightgray}{RGB}{166, 166, 166}

\def\BibTeX{{\rm B\kern-.05em{\sc i\kern-.025em b}\kern-.08em
    T\kern-.1667em\lower.7ex\hbox{E}\kern-.125emX}}
\begin{document}

\title{TUGS: Physics-based Compact Representation of Underwater Scenes by Tensorized Gaussians}


\author{
    \IEEEauthorblockN{Shijie Lian\textsuperscript{\rm1,2 *}, Ziyi Zhang\textsuperscript{\rm3,5 *}, Hua Li\textsuperscript{\rm5 \Letter}, Laurence Tianruo Yang\textsuperscript{\rm1,4 \Letter}, Mengyu Ren\textsuperscript{\rm5}, Debin Liu\textsuperscript{\rm4}, Wenhui Wu\textsuperscript{\rm6}}
    \IEEEauthorblockA{\textsuperscript{\rm1}Huazhong University of Science and Technology \quad \textsuperscript{\rm2}Zhongguancun Academy}
    \IEEEauthorblockA{\textsuperscript{\rm3}The Chinese University of Hong Kong, Shenzhen \quad \textsuperscript{\rm4}Zhengzhou University \quad \textsuperscript{\rm5}Hainan University \quad \textsuperscript{\rm6}Shenzhen University}
    \thanks{$^{*}$ Equal contribution.}
    \thanks{\Letter~Corresponding authors: Hua Li (\texttt{lihua@hainanu.edu.cn}) and Laurence Tianruo Yang (\texttt{LTYANG@zzu.edu.cn}).}
}
\makeatletter
\def\footnoterule{\kern-3\p@\hrule\@width.4\columnwidth\kern2.6\p@}
\makeatother

\maketitle

\begin{abstract}

Underwater 3D scene reconstruction is crucial for multimedia applications in adverse environments, such as underwater robotic perception and navigation.
However, the complexity of interactions between light propagation, water medium, and object surfaces poses significant difficulties for existing methods in accurately simulating their interplay.
Additionally, expensive training and rendering costs limit their practical application.
Therefore, we propose Tensorized Underwater Gaussian Splatting (\methodName), a compact underwater 3D representation based on physical modeling of complex underwater light fields.
\methodName~ includes a physics-based underwater Adaptive Medium Estimation (AME) module, enabling accurate simulation of both light attenuation and backscatter effects in underwater environments, and introduces Tensorized Densification Strategies (TDS) to efficiently refine the tensorized representation during optimization.
\methodName~is able to render high-quality underwater images with faster rendering speeds and less memory usage.
Extensive experiments on real-world underwater datasets have demonstrated that \methodName~can efficiently achieve superior reconstruction quality using a limited number of parameters. The code is available at \url{https://liamlian0727.github.io/TUGS}
\end{abstract}
  
\vspace{2mm}
\section{Introduction}
\label{sec:intro}
Underwater 3D scene reconstruction is an essential part of underwater robotics perception and navigation in adverse environments, enabling multimedia applications such as SLAM \cite{macario2022comprehensive}, large-scale scene reconstruction \cite{ gu2024ue4, tancik2022block}, and scene understanding \cite{chen2023clip2scene, qin2024langsplat, peng2023openscene}.
However, Underwater 3D reconstruction is challenging due to the different properties of the medium compared to air. 
Specifically, the absorption and backscatter of light in water cause significant attenuation and degradation of visual content in underwater environments \cite{SeaThru_2018_CVPR, akkaynak2017space}, causing inaccuracies in color and density estimation. 

In recent years, Neural Radiance Field (NeRF) \cite{NeRF_2020_ECCV}, as a new implicit representation for 3D reconstruction, enables high-quality novel view synthesis. 
However, the formulations of the original NeRF \cite{NeRF_2020_ECCV}~and its follow-up variants \cite{MipNeRF_2022_CVPR} assume that images were acquired in clear air and the rendered image is composed solely of the object radiance.
Specifically, the assumption of zero density between the camera and the object does not consider the absorption and scattering of light produced by different media. 
To address this issue, SeaThru-NeRF\cite{SeaThruNeRF_2023_CVPR} incorporates the water medium into the rendering model by using two radiance fields: one for the geometry and another for the water medium.
However, the high training and rendering costs of SeaThru-NeRF make it challenging to synthesize high-quality images in real-time, thus limiting its practical application in underwater devices.

\begin{figure}[t]
  \centering
  \includegraphics[width=1\linewidth]{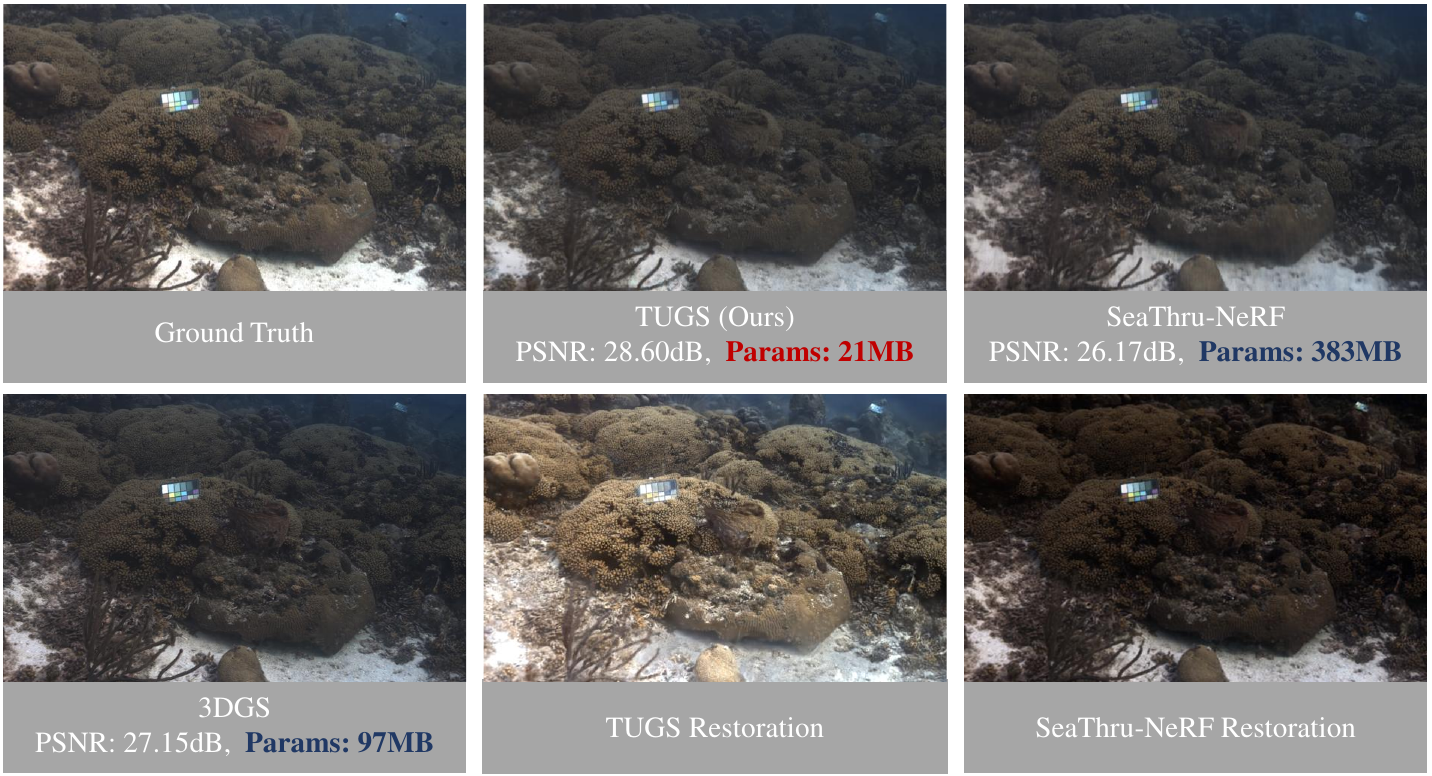}
  \vspace{-7mm}
  \caption{SeaThru-NeRF \cite{SeaThruNeRF_2023_CVPR}, 3DGS \cite{3DGS_2023_TOG}, and the proposed method \methodName~are trained on the Cura\c{c}ao scene of SeaThru-NeRF dataset. It can be seen that our model achieves the best PNSR with a size of only \textbf{21 MB}.}
  \vspace{-5.5mm}
  \label{fig:simple_comp}
\end{figure}

More recently, 3D Gaussian Splatting (3DGS) \cite{3DGS_2023_TOG} introduces an explicit point-based scene representation with learnable 3D Gaussians (e.g., position, covariance, color, and opacity), enabling rapid optimization and real-time rendering for clear-medium scenarios. 
However, when directly applied to underwater imagery, 3DGS often allocates many low-opacity primitives to explain medium-induced attenuation and scattering, which can introduce redundancy, artifacts, and increased storage and rendering costs in complex underwater environments. 
Recent 3DGS-based underwater methods, such as SeaSplat \cite{seasplat_2024_arXiv} and UW-GS \cite{uwgs_2025_WACV}, incorporate physically motivated image-formation cues to better account for the medium; nevertheless, they largely retain the original primitive-level parameterization and densification, and rely on additional learnable components or auxiliary signals (e.g., appearance networks, pseudo-depth, motion masks), which increases complexity and storage.

To address the above issues, we propose the Tensorized Underwater Gaussian Splatting (\methodName), which utilizes different mode-1 slices of tensorized Gaussians to simulate the underwater object and medium.
\methodName~is a compact representation based on advanced tensor decomposition, which reduces the number of parameters required for simulating objects and water media.
Specifically, we couple the object and medium as two mode-1 slices of $\mathcal{G}\!\in\!\mathbb{R}^{2\times N\times M}$ and enforce factor sharing via a rank-$R$ CP decomposition, making the medium branch an $O(R)$ add-on and reducing parameters from $2MN$ to $(M+N+2)R$.
As shown in \cref{fig:simple_comp}, compared to 3DGS, TUGS uses only 20\% of the parameters while maintaining a similar rendering speed. Meanwhile, SeaThru-NeRF\cite{SeaThruNeRF_2023_CVPR} requires 383 MB and renders at 0.09 FPS, whereas our method achieves 106 FPS with just 21 MB.
This efficiency makes \methodName~particularly suitable for memory-constrained multimedia applications in adverse environments, such as underwater robotics and vehicles.

Moreover, we propose an Adaptive Medium Estimation (AME) module to simulate the imaging process in underwater environments, preventing the generation of a large number of redundant Gaussians used to represent the water medium. Based on the physics of underwater imaging, the AME module models the impact of underwater light attenuation and backscatter on objects.
Specifically, AME turns the medium slice into a depth-aware attenuation/backscatter estimate and composes it with the restored object slice through a physics-based underwater image formation model.
Compared to other 3DGS-based underwater 3D reconstruction methods\cite{seasplat_2024_arXiv,uwgs_2025_WACV}, \methodName~does not require the generation of additional Gaussian primitives or the introduction of MLPs to adapt to the effects caused by the water medium, further reducing the amount of memory consumed.
Additionally, we propose a companion optimization strategy for \methodName, which consists of a Tensorized Densification Strategy to reduce the computational overhead of the model in the densification process, and a well-designed underwater reconstruction loss function to help the model reconstruct the underwater scene accurately and efficiently.

To validate our method, we evaluate it on the established benchmark underwater dataset, SeaThru-NeRF \cite{SeaThruNeRF_2023_CVPR} and the simulated dataset \cite{MipNeRF_2022_CVPR}.
The results of our evaluation demonstrate the effectiveness of the proposed method in achieving high-quality and efficient underwater reconstruction while maintaining low storage costs.
Furthermore, due to the characteristics of Tensorized Gaussians, our approach allows for a tunable trade-off between rendering quality and storage usage,  making it better aligned with the requirements of multimedia applications.
Visualization results show that after removing the light attenuation and backscatter from our rendering, our model can restore the realistic colors of underwater objects and produce reasonable visual effects.
In summary, we make the following contributions:

\begin{enumerate}
    \item We propose a compact underwater scene representation, Tensorized Underwater Gaussian Splatting (\methodName), which effectively compresses storage and simulates light field variations induced by the medium through the introduction of higher-order tensors.
    \item We designed the Adaptive Medium Estimation (AME) module and Tensorized Densification Strategies (TDS) according to the physics-based underwater image formation model, where AME explicitly estimates attenuation/backscatter for image composition and TDS performs efficient cloning/pruning directly on the CP factors to preserve compactness during optimization.
    \item Extensive experiments on the real-world and simulated datasets demonstrate that, compared to other methods, our approach achieves high-quality underwater reconstruction with \textbf{2--8$\times$} smaller storage. 
\end{enumerate}


\section{Related work}
\label{sec:related work}

\subsection{3D Scene Reconstruction}
3D scene reconstruction from multi-view images is a fundamental task for multimedia applications. Recently, considerable attention has been paid to reconstructing 3D scenes, particularly with approaches like NeRF \cite{NeRF_2020_ECCV} based on implicit representation and 3DGS \cite{3DGS_2023_TOG} that combine explicit and implicit representation. NeRFs \cite{SeaThruNeRF_2023_CVPR, NeRF_2020_ECCV, MipNeRF_2022_CVPR} model the 3D scene as a radiance field and synthesize novel views through volume rendering. Currently, NeRF has made significant advancements in various domains, such as large-scale scene reconstruction \cite{gu2024ue4, tancik2022block}, dynamic scene reconstruction \cite{du2021nerflow}, and underwater scene reconstruction \cite{SeaThruNeRF_2023_CVPR}.
SeaThru-NeRF \cite{SeaThruNeRF_2023_CVPR} is based on the SeaThru \cite{SeaThru_2018_CVPR} image formation model using multiple MLPs to model objects and the medium separately to address the issue of how the medium affects the appearance of objects in underwater or foggy scenes. However, due to the dense sampling and querying of NeRF's volume rendering within multiple MLPs, the training and rendering costs are too extensive to achieve real-time rendering. In contrast, 3DGS \cite{3DGS_2023_TOG} represents the scene as a set of Gaussian points and achieves real-time rendering through rasterization, demonstrating strong applicability in multimedia applications like autonomous navigation \cite{zhou2024drivinggaussian,matsuki2024gaussian} and 3d scene understanding \cite{qin2024langsplat}.
Despite these impressive advances, efficient 3D reconstruction in adverse scenes remains a challenge. 

\subsection{Image Restoration in Scattering Media}
In underwater environments where medium parameters exhibit strong wavelength dependency, the SeaThru model \cite{SeaThru_2018_CVPR, DeepSeeColor_2023_ICRA, akkaynak2017space} was introduced to address these wavelength-specific effects. 
To be specific, image formation in fog, haze, or underwater differs from clear air due to two key factors: (1) the direct signal from the object is attenuated based on distance and wavelength, and (2) backscatter which is independent of the scene content but intensifies with distance, reducing visibility and contrast while distorting colors. These processes can be modeled as the following:
\begin{equation}
    \label{eq:seathru}
    I_c\left (i,j \right) = A_c\left (i,j \right) \cdot J_c\left (i,j \right) + B_c\left (i,j \right),
\end{equation}
where the light intensity in the channel $c \in \left \{R, G, B \right\}$~of the target along the light ray to the camera image sensor pixel $(i, j)$~is denoted as $J_c\left (i,j \right)$~and the light intensity measured by the sensor at that pixel is denoted as $I_c\left (i,j \right)$.
$A_c\left (i,j \right)$~and $B_c\left (i,j \right)$~represent wavelength-dependent light attenuation and backscatter, respectively.
\section{Tensorized Underwater Gaussian Splatting}
\label{sec:method}

\begin{figure*}[t]
  \centering
  \includegraphics[width=0.98\linewidth]{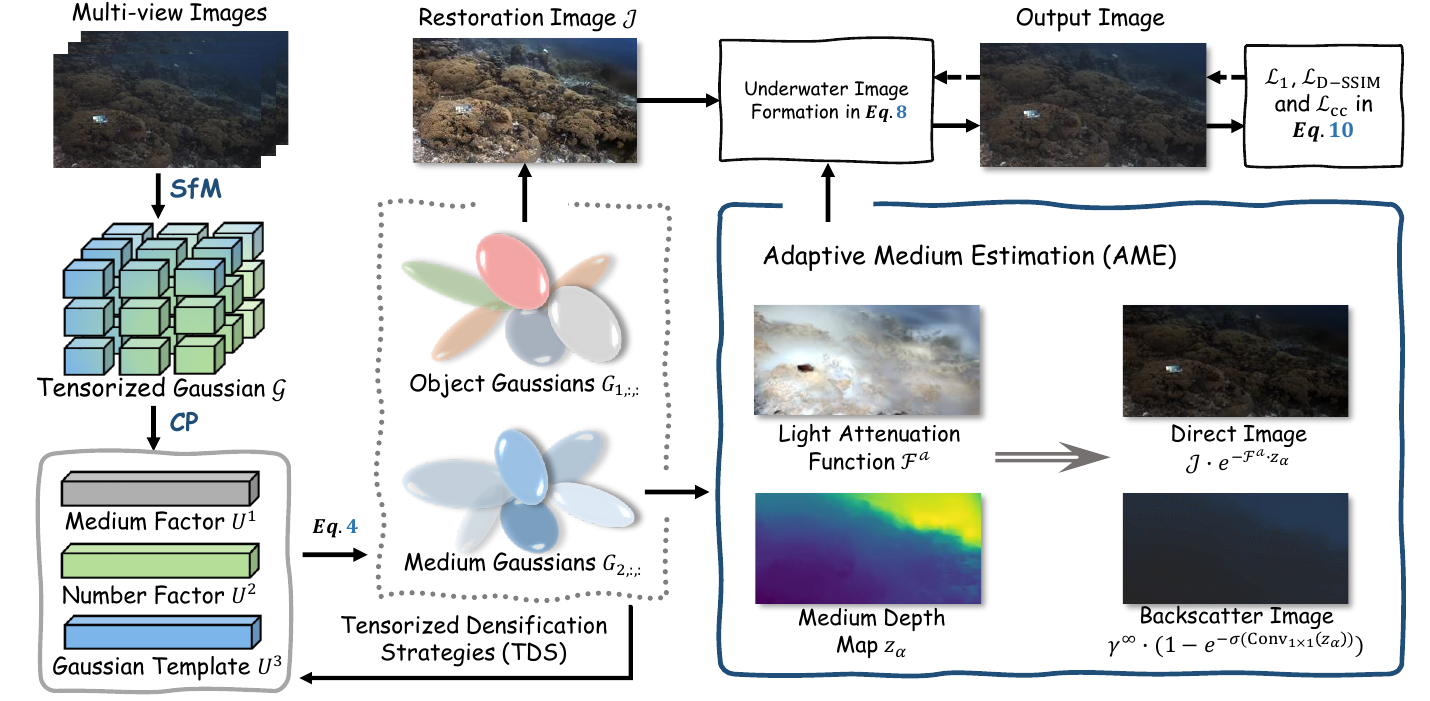}
  \vspace{-3mm}
  \caption{TUGS models the underwater object and the water medium by using different mode-1 slices of a high-order tensorized Gaussian $\mathcal{G}$, and simultaneously represents them using the mode factors $[\mathbf{U}^1 \text{, }\mathbf{U}^2 \text{, }\mathbf{U}^3]$, reducing the parameter count by approximately 60-85\% compared to 3DGS \cite{3DGS_2023_TOG} (in \cref{sub:TensorizedGaussians}). When synthesizing images, TUGS renders the medium Gaussian as a light attenuation and backscatter image through the Adaptive Medium Estimation (AME) model and blends it with the restoration image from the object Gaussian through the underwater image formation model for the final output (in \cref{sub:AdaptiveMediumEstimation}).
  \textbf{CP}~is the CP decomposition \cite{CP_C_1970, CP_P_1970}~and \textbf{TDS}~stand for Tensorized Densification Strategies in \cref{sub:tds}. 
  }
  \vspace{-5mm}
  \label{fig:main_framework}
\end{figure*}

\subsection{Tensorized Gaussians}
\label{sub:TensorizedGaussians}

Due to the complex and adverse of underwater environments, achieving realistic 3D reconstruction requires a representation that accounts for both the object signal and water-medium effects. Specifically, underwater image formation involves range-dependent light attenuation and depth-dependent backscatter that severely degrade geometry and appearance recovery. Following SeaThru \cite{SeaThru_2018_CVPR}, we model underwater scenes with attenuation and backscatter:
\begin{equation}
    \label{eq:a_in_underwater}
    A_c\left (i,j \right) = e^{-f^{a}_c\left (i,j \right) \cdot z\left (i,j \right)}\text{,}
\end{equation}
\begin{equation}
    \label{eq:b_in_underwater}
    B_c \left (i,j \right) = \gamma^{\infty}_c \cdot \left (1 - e^{-\beta^B_c\cdot z\left (i,j \right)}\right)\text{,}
\end{equation}
where $\gamma^{\infty}_c \in \mathbb{R}_{\ge0}$ denotes the backscatter color at infinite distance, $\beta^B_c\in \mathbb{R}_{\ge0}$ controls backscatter strength, and $f^{a}_c\left (i,j \right)$ is a per-pixel attenuation function dependent on depth, reflectance, ambient light, and water scattering properties.
Therefore, efficient real-time reconstruction requires a compact representation that jointly models object appearance/geometry and water-medium effects as defined by \cref{eq:a_in_underwater,eq:b_in_underwater}. 



As shown in \cref{fig:main_framework}, TUGS represents the scene with a tensorized Gaussian $\mathcal{G}\in\mathbb{R}^{2\times N\times M}$, where the two mode-1 slices correspond to the object component and the medium component. We denote the resulting slices as $\mathbf{G}_{1,:,:}$ and $\mathbf{G}_{2,:,:}$, which are rasterized to produce the restoration image $\mathcal{J}$ and the attenuation-related field $\mathcal{F}^a$, respectively; these outputs are then combined by the AME module in \cref{sub:AdaptiveMediumEstimation} through a physically-based formation model.

To avoid using two independent Gaussian sets (with parameter size $2MN$), we apply a rank-$R$ CANDECOMP/PARAFAC (CP) decomposition \cite{CP_C_1970} to $\mathcal{G}$ and optimize only three shared factor matrices: the medium factor $\mathbf{U}^1\in\mathbb{R}^{2\times R}$, the number factor $\mathbf{U}^2\in\mathbb{R}^{N\times R}$, and the Gaussian template $\mathbf{U}^3\in\mathbb{R}^{M\times R}$. Specifically,
\begin{equation}
    \label{eq:decp}
    \mathcal{G} \approx \sum^R_{r=1} \mathbf{u}^1_{:,r} \circ \mathbf{u}^2_{:,r} \circ \mathbf{u}^3_{:,r}\text{,}
\end{equation}
where $\circ$ denotes the vector outer product and $\mathbf{u}^m_{:,r}$ is the $r$-th column of $\mathbf{U}^m$. Sharing $\mathbf{U}^2$ and $\mathbf{U}^3$ across both slices imposes a low-rank prior that regularizes the object and medium components to be consistent, while reducing the learnable parameters to $\left(M+N+2\right)\times R$; equivalently, introducing the additional medium slice costs only $R$ parameters in $\mathbf{U}^1$ rather than a full extra $MN$ set. For instance, when $R=20$, $N=2\times10^5$, and $M=59$, \methodName\ learns $\left(M + N + 2\right)\times R \approx 4\times10^6$ parameters, compared to $M\times N = 12\times10^6$ in the original 3DGS representation, thus reducing the total number of learnable parameters by about 66\%.

\subsection{Adaptive Medium Estimation} 
\label{sub:AdaptiveMediumEstimation}
In the previous subsection, we obtained the clean underwater image $\mathcal{J}$~with the light attenuation function $\mathcal{F}^a$.
Then, we'll blend them into the final output through the Adaptive Medium Estimation (AME) module.
%
Specifically, for a given camera direction $d$, $\mathcal{J}$~and $\mathcal{F^a}$~with depth $\mathbf{z}$~are rendered as:
\begin{equation}
    \label{eq:ras_obj}
    \mathcal{J}\text{, }\mathbf{z_u} = r\left(\mathbf{G}_{1,:,:},d\right),
\end{equation}
\begin{equation}
    \label{eq:ras_att}
    \mathcal{F}^a\text{, }\mathbf{z_a} = r\left(\mathbf{G}_{2,:,:},d\right),
\end{equation}
where $\mathcal{J} \text{, } \mathcal{F}^a \in \mathbb{R}^{H\times W\times 3}$~and $\mathbf{z_u} \text{, } \mathbf{z_a} \in \mathbb{R}^{H\times W}$, $r{\left (\mathbf{G},d \right)}$~is the differentiable rasterization of Gaussian $\mathbf{G}$~at camera direction $d$. 
Subsequently, we estimate the backscatter image $\mathcal{B}\in \mathbb{R}^{H\times W\times 3}$~based on \cref{eq:b_in_underwater}, and the backscatter coefficient $\bm{\beta^B}$~is modeled using a $ 1 \times 1 $~convolution layer $\mathrm{Conv}_{1\times 1}\left (\cdot\right)$, where the layer's weights are denoted as $ \mathcal{W}\in \mathbb{R}^{3\times1\times1\times1} $. Consequently, $\mathcal{B}$~is expressed as:
\begin{equation}
    \label{eq:b}
    \mathcal{B}=\bm{\gamma^{\infty}} \cdot \left ( 1 - e^{-\sigma\left (\mathrm{Conv}_{1\times 1} \left(\mathcal{\mathbf{z_a}}\right)\right)}\right),
\end{equation}
where $\bm{\gamma^{\infty}}$~is a non-negative learnable variable and $\sigma\left (\cdot\right)$ denotes the ReLU activation function. Then, the image $\mathbf{I} \in \mathbb{R}^{H \times W \times 3}$, synthesized based on the underwater image formation model, can be rendered as follows:
\begin{equation}
    \label{eq:udfm_use}
    \mathbf{I}= \mathcal{J}\cdot e^{-\mathcal{F}^a \cdot \mathbf{z_a}} + \mathcal{B}\text{.}
\end{equation}
In addition, in order to make the Gaussian primitive that renders the light attenuation more focused on the information related to the scattering of the water body, we use $\mathbf{z_a}$~instead of $\mathbf{z_u}$~in \cref{eq:b}~and \cref{eq:udfm_use}.

\begin{table*}[t]
    \centering
     \caption{Quantitative evaluation on the SeaThru-NeRF dataset, where $R$ stand for rank in mode factors $[\mathbf{U}^1 \text{, }\mathbf{U}^2 \text{, }\mathbf{U}^3]$. We show the PSNR$\uparrow$, SSIM$\uparrow$, and LPIPS$\downarrow$. Light red and light yellow correspond to the first and second. }
     \vspace{-2mm}
    \label{tab:3dgs.comp}
    \renewcommand{\arraystretch}{1.14}
    \setlength{\tabcolsep}{0.8mm}{
    \scalebox{1}{\begin{tabular}{c|cccc|cccc|cccc|cccc|c}
    \hline
     \hline
    \multirow{2}{*}{Method} &\multicolumn{4}{c|}{IUI3 Red Sea} &\multicolumn{4}{c|}{Cura\c{c}ao} &\multicolumn{4}{c|}{J.G. Red Sea}&\multicolumn{4}{c|}{Panama} & Avg.\\ 
    \cline{2-17}
     &  PSNR& SSIM & LPIPS & Storage& PSNR& SSIM & LPIPS & Storage& PSNR& SSIM & LPIPS & Storage  &PSNR& SSIM & LPIPS& Storage & FPS\\
    \hline
    SeaThru-NeRF \cite{SeaThruNeRF_2023_CVPR}& 25.84 & 0.85 & 0.30 & 383 MB & 26.17& 0.81 & 0.28 & 383 MB & 21.09 & 0.76 & 0.29 & 383 MB & 27.04 & 0.85 & 0.22 & 383 MB & 0.09\\
    TensoRF \cite{tensorf_2022_ECCV} & 17.33 & 0.55 & 0.63 & 66 MB & 23.38 & 0.79 & 0.45 & 66 MB & 15.19 & 0.51 & 0.59 & 66 MB & 20.76 & 0.75 & 0.38 & 66 MB & 0.21\\
    3DGS \cite{3DGS_2023_TOG}& 28.28 & \cellcolor{lightyellow}0.86 & \cellcolor{lightyellow}0.26 & 105 MB & 27.15 & 0.85 & 0.25 & 97 MB & 20.26 & 0.82 & \cellcolor{lightyellow}0.23 & 89 MB & 30.23 & 0.89 & 0.19 & 75 MB & 154.9\\
    SeaSplat \cite{seasplat_2024_arXiv}& 26.47 & 0.85 & 0.28 & 140 MB& 27.79& 0.86 & 0.27 & 129 MB & 19.21 & 0.70 & 0.35 & 138 MB & 29.79 & 0.88 & 0.19 & 104 MB& 80.7\\
    UW-GS \cite{uwgs_2025_WACV}& 27.06 & 0.84 & 0.27 & 78 MB& 27.62 &\cellcolor{lightyellow} 0.87 & 0.25 & 64 MB & 20.05 & 0.71 & 0.32 & 68 MB & \cellcolor{lightyellow}31.36 & \cellcolor{lightyellow}0.90 & 0.17 & 61 MB & 78.6\\
    \hline
    \methodName~($R=20$) & \cellcolor{lightyellow}28.98 & \cellcolor{lightyellow}0.86 & \cellcolor{lightyellow}0.26 &  
    \cellcolor{lightred} 18 MB &
    \cellcolor{lightyellow}28.60 & \cellcolor{lightred}0.88 & \cellcolor{lightyellow}0.23 &
    \cellcolor{lightred} 21 MB &
    \cellcolor{lightyellow}22.21 & \cellcolor{lightyellow}0.84 & \cellcolor{lightyellow}0.23 & 
    \cellcolor{lightred} 11 MB & 31.19 & \cellcolor{lightred}0.92 & \cellcolor{lightyellow}0.16 &
    \cellcolor{lightred} 12 MB & 106.7\\
    \methodName~($R=30$)& \cellcolor{lightred}29.36 & \cellcolor{lightred}0.87 & \cellcolor{lightred}0.25 & 
    \cellcolor{lightyellow}31 MB & 
    \cellcolor{lightred}28.71 & \cellcolor{lightyellow}0.87 & \cellcolor{lightred}0.22 &
    \cellcolor{lightyellow}43 MB & 
    \cellcolor{lightred}22.43 & \cellcolor{lightred}0.86 & \cellcolor{lightred}0.22 & 
    \cellcolor{lightyellow}37 MB & 
    \cellcolor{lightred}31.51 & \cellcolor{lightred}0.92 & \cellcolor{lightred}0.15 & 
    \cellcolor{lightyellow}34 MB & 82.1\\
    \hline \hline
    \end{tabular}}}
    \vspace{-5mm}
\end{table*}

\subsection{Tensorized Densification Strategies}

\label{sub:tds}

With the proposed tensorized representation, using only the Structure-from-Motion (SfM) initialization (i.e., without any densification) already yields an extremely compact model, typically around \textbf{1--3\,MB} in our settings. However, SfM initialization can be sparse or unreliable in challenging underwater conditions, which may lead to missing structures and suboptimal reconstruction quality. To improve scene coverage while preserving compactness, we adapt the densification strategy of 3DGS \cite{3DGS_2023_TOG} to our factorized formulation.

In 3DGS, densification dynamically clones/splits Gaussians and prunes low-contribution primitives. Directly applying splitting in our setting would require repeatedly materializing $\mathcal{G}$ and re-running tensor operations, introducing unnecessary overhead. Therefore, we perform densification directly on the mode factors, consisting of: (i) gradient-based cloning, where for primitives with gradient magnitude above a threshold $t_g$ we append a copy of the corresponding row in $\mathbf{U}^2$ (adding only $R$ parameters for a new primitive); (ii) opacity-based pruning, where we remove primitives whose opacity is below $t_o$; and (iii) periodic opacity reset, where we set the opacity row in the template to zero to encourage re-allocation. Specifically, we retain primitives by:
\begin{equation}
    \label{eq:remove_gs}
    \mathbf{U^2_{retain}}=\left\{\mathbf{u}_{i,:}^2 \,\middle|\, \mathbf{u}_{i,:}^2{\left( \mathbf{u}_{o,:}^3\odot\mathbf{u}_{1,:}^1\right)}^T\ge t_o\right\}\text{,}
\end{equation}
where $\odot$ denotes the Khatri--Rao product and $\mathbf{u}_{o,:}^3$ is the row of $\mathbf{U}^3$ corresponding to opacity. This design avoids repeated tensor reconstruction and keeps the densification process efficient under the CP-factorized representation.

\subsection{Loss Function}

Due to the lack of a medium-free reference image that fully removes underwater effects, the restoration image $\mathcal{J}$ cannot be directly supervised. 
To mitigate this issue, during 3DGS-style optimization, besides the standard $L_1$ reconstruction loss and the D-SSIM loss, we follow underwater color correction works \cite{SeaThru_2018_CVPR, DeepSeeColor_2023_ICRA} and introduce a color correction loss that encourages balanced per-channel statistics in $\mathcal{J}$:
\begin{equation}
    \label{eq:colorloss}
    \mathcal{L}_\mathrm{cc}=\sum_{c}\left(\left(m\left(\bm{J_c}\right)-0.5\right)^2 + \left(s\left(\bm{J_c}\right)-s(\bm{D_c})\right)^2\right)\text{,}
\end{equation}
where $c\in\{R,G,B\}$, $m(\cdot)$ and $s(\cdot)$ compute the mean and standard deviation, and $\mathcal{D}=\mathcal{I}^{gt}-\mathcal{B}$ denotes the estimated direct signal. 
In practice, $\mathcal{L}_\mathrm{cc}$ acts as a soft white-balance regularizer that reduces color cast and prevents overly dark restorations, improving visual clarity under varying water qualities and illumination conditions.

\section{Experiments}

\subsection{Implementation and Optimization}
For initialization, we first initialize the object slice $\mathbf{G}_{1,:,:}$ using the standard 3DGS pipeline from the underwater input images, and initialize the medium slice $\mathbf{G}_{2,:,:}$ by copying the corresponding factors for a lightweight warm start. We train TUGS on an RTX 4090 GPU for 20,000 steps using Adam, with learning rates and optimizer hyperparameters kept consistent with 3DGS \cite{3DGS_2023_TOG}.
For AME, the learnable parameter $\lambda^{\infty}$ is initialized as $[0.1,0.2,0.3]$, and the convolutional layer in \cref{eq:b} uses uniform initialization; both are optimized with Adam at an initial learning rate of $10^{-3}$. For Tensorized Densification Strategies (TDS), we set $t_g=10^{-3}$ for copying Gaussians and $t_o=0.1$ for pruning low-opacity Gaussians, and reset opacity to zero every 1,000 steps. TDS is applied during the first 10,000 steps to ensure sufficient primitives for stable reconstruction.

\begin{figure*}[t]
  \centering
  \includegraphics[width=1\linewidth]{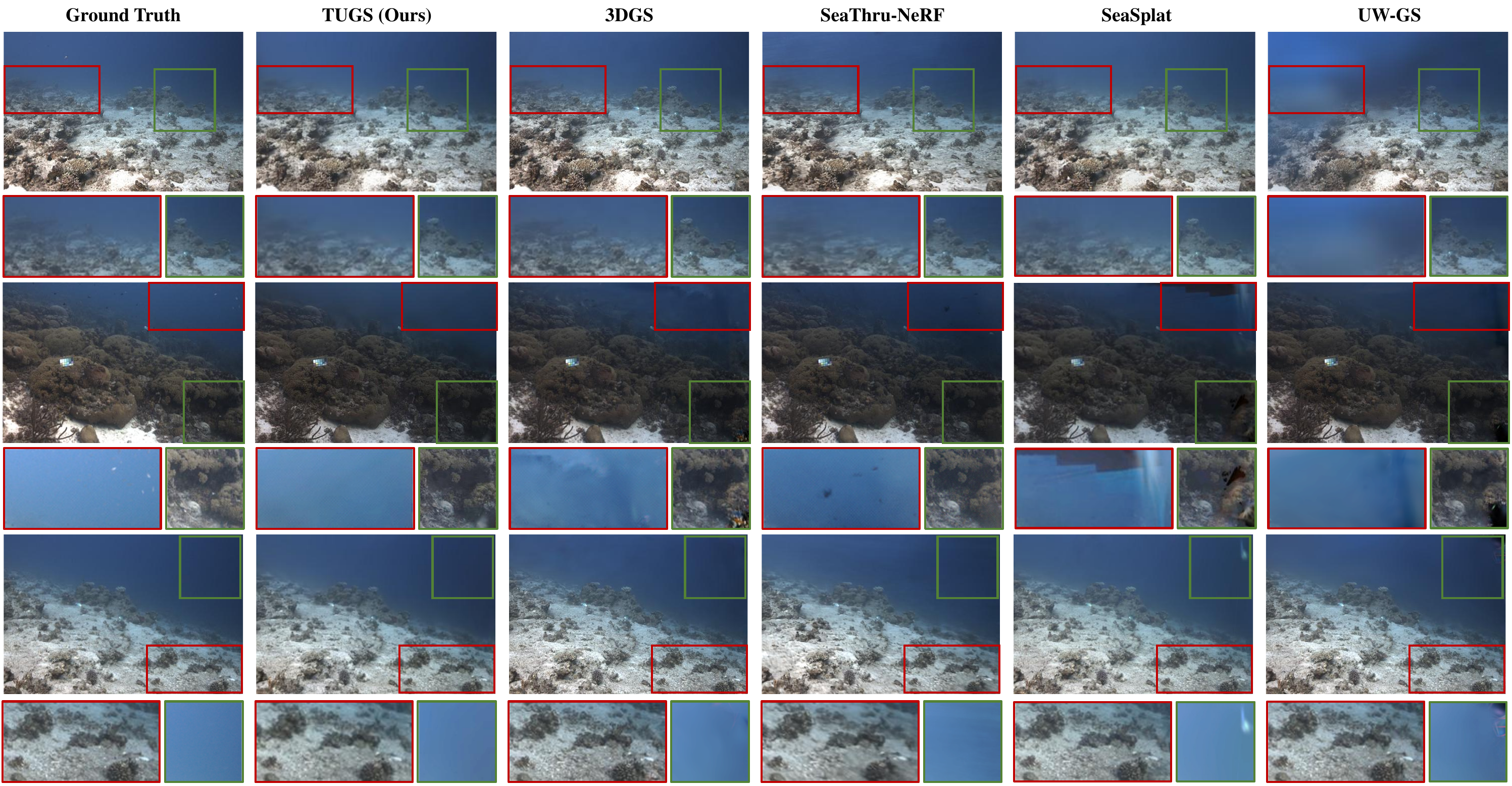}
  \vspace{-7mm}
  \caption{Novel view rendering comparisons in SeaThru-NeRF dataset \cite{SeaThruNeRF_2023_CVPR}. We adjusted the image brightness of two highlighted regions in row 4 and the green highlighted region in row 6 with the same settings for more visual comparisons of the different methods.
  }
  \vspace{-2mm}
  \label{fig:comp}
\end{figure*}

\begin{figure}[t]
  \centering
  \includegraphics[width=1\linewidth]{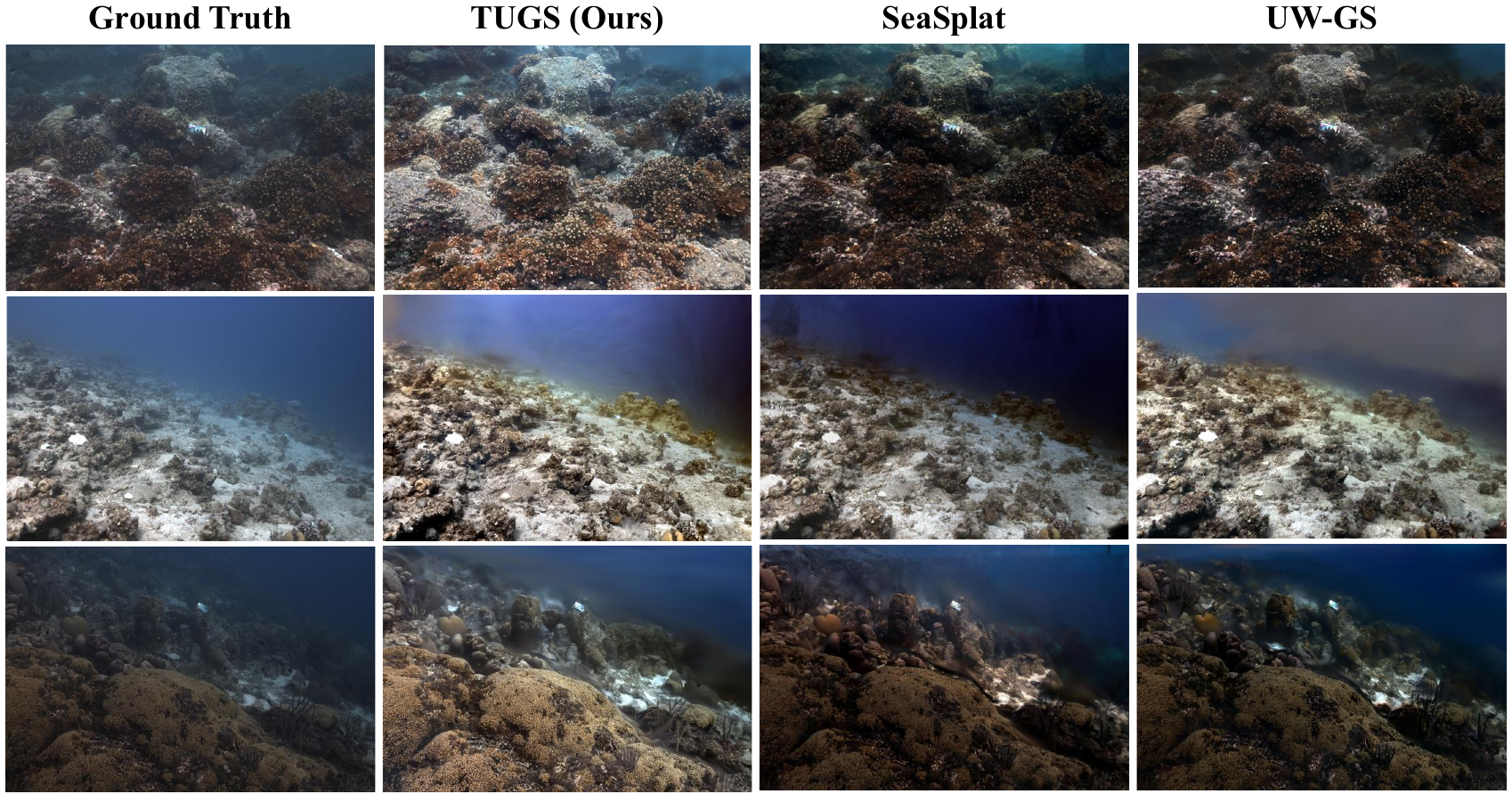}
  \vspace{-7mm}
  \caption{Novel view synthesis without water media in the SeaThru NeRF dataset \cite{SeaThruNeRF_2023_CVPR}. TUGS restores the underlying colors of the scene more vividly, and foreground details are clearer.}
  \vspace{-5mm}
  \label{fig:en_comp}
\end{figure}

\subsection{Real-world and Simulated Datasets}
We evaluate on the SeaThru-NeRF dataset \cite{SeaThruNeRF_2023_CVPR}, which contains four real-world underwater scenes: IUI3 Red Sea, Cura\c{c}ao, Japanese Garden Red Sea, and Panama. 
To further test the performance of our method, we simulate fog environments on the Mip-NeRF 360 dataset \cite{MipNeRF_2022_CVPR}, following \cite{SeaThruNeRF_2023_CVPR}. 


\subsection{Baseline Methods and Evaluation Metrics}
To ensure fairness, the input to all methods is the same set of white-balanced linear images.
For scenes with the medium, we compare the SeaThru-NeRF \cite{SeaThruNeRF_2023_CVPR}, TensoRF \cite{tensorf_2022_ECCV}, 3DGS \cite{3DGS_2023_TOG}, SeaSplat \cite{seasplat_2024_arXiv}, and UW-GS \cite{uwgs_2025_WACV}.
As shown in \cref{tab:3dgs.comp}, we quantitatively analyze the rendering results across different methods by calculating PSNR, SSIM, and LPIPS, and Storage (the parameters needed to represent the same scene).
For reconstructing clean medium-free images, since we do not have ground truth images without the medium, we follow the setup of SeaThru \cite{SeaThru_2018_CVPR}~and show the results of the visual comparison with the SeaSplat and UW-GS in \cref{fig:en_comp}.


\subsection{Results and Analysis}

\cref{tab:3dgs.comp} shows that \methodName~requires only \textbf{13-45\% }of the number of parameters of 3DGS to achieve top performance in most scenes and metrics.
Even compared with SeaThru-NeRF, which has about 10-20 times the number of parameters, we can remain competitive across all metrics.
For TensoRF \cite{tensorf_2022_ECCV}, though also tensor-decomposed, produces over-smoothed renderings and underperforms, as it does not explicitly model the relationship between the underwater object and the medium.

\begin{figure}[t]
  \centering
  \includegraphics[width=0.98\linewidth]{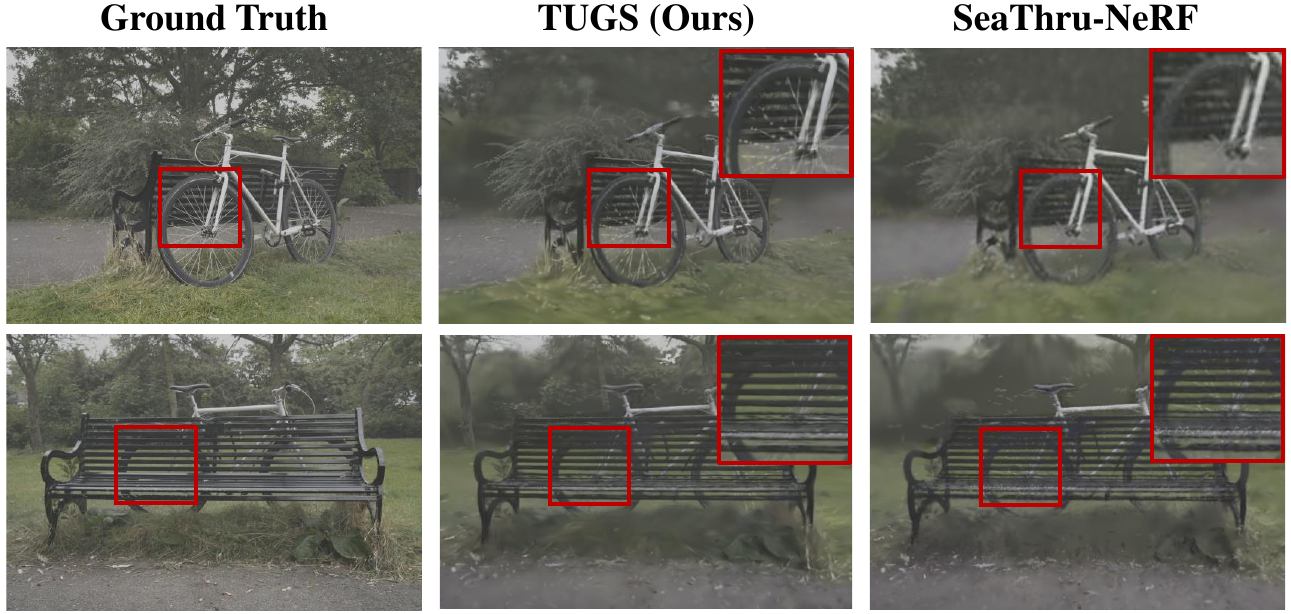}
  \vspace{-2mm}
  \caption{Synthesizing novel views in a foggy environment.}
  \vspace{-3mm}
  \label{fig:fog_comp}
\end{figure}

Additionally, as shown in the red-highlighted areas of the first row in \cref{fig:comp}, UW-GS \cite{uwgs_2025_WACV}, a Gaussian Splatting method specifically designed for underwater environments, exhibits noticeable artifacts when rendering water.
In the third row, the red-highlighted area reveals unnatural gaps in the background of the image synthesized by SeaSplat \cite{seasplat_2024_arXiv}. 
In contrast, our method utilizes the AME module to properly blend the restoration image, light attenuation, and backscatter, producing the final output with visually satisfactory results.
In addition, as shown in \cref{fig:en_comp}, our medium-free reconstructions recover more vivid underlying scene colors and clearer foreground details compared to SeaSplat and UW-GS.
Moreover, it can be seen in \cref{tab:fog} and \cref{fig:fog_comp}~that the same advantages of our method are observed in the simulated dataset. It outperforms SeaThru-NeRF in fogged images and all types of metrics.

\begin{table}[!t]
    \centering
     \vspace{-3mm}
     \caption{Quantitative evaluation on the simulated dataset.}
     \vspace{-3mm}
    \renewcommand{\arraystretch}{1}
    \setlength{\tabcolsep}{=3mm}
    {\scalebox{1}{\begin{tabular}{c|cccc}
    \hline\hline
    Method & PSNR & SSIM & LPIPS & Storage\\ 
    \hline 
    SeaThru-NeRF \cite{SeaThruNeRF_2023_CVPR}& \cellcolor{lightyellow}20.07 & \cellcolor{lightyellow}0.66  & \cellcolor{lightyellow}0.61 & 383 MB\\
    3DGS \cite{3DGS_2023_TOG}& 18.94 & 0.61 & 0.69 & \cellcolor{lightyellow}135 MB\\
    TUGS ($R=30$) & \cellcolor{lightred}23.64 & \cellcolor{lightred}0.74 & \cellcolor{lightred}0.53 & \cellcolor{lightred}58 MB\\
    \hline  \hline
    \end{tabular}}}
    \vspace{-3mm}\label{tab:fog}
\end{table}

\subsection{Ablations}

We decompose the components of \methodName~and validate the effectiveness of each module through incremental additions. 
All ablation experiments are conducted on the IUI3 Red Sea scene from the SeaThru-NeRF dataset \cite{SeaThruNeRF_2023_CVPR}, with the CP decomposition rank set to 30. TG denotes tensorized Gaussian, AME denotes adaptive medium estimation, Loss is our loss function, and TDS is our tensorized densification strategy.
As shown in \cref{tab:ablation}, initially, when we only apply CP decomposition to the parameters of 3DGS and remove the original densification, the performance degrades compared to 3DGS. 
However, when we introduce our AME module and loss function, the model demonstrates a significant performance improvement. 
Notably, without the densification strategy, the model $\text{TG + AME + Loss}$~typically contains around 20,000 Gaussian primitives (approximately \textbf{3MB}), which results in exceptionally fast training speed and minimal memory requirements. This configuration is well-suited for deployment on resource-constrained underwater edge devices. Furthermore, after incorporating our TDS, the model's performance improves considerably, enabling precise underwater scene modeling with only about 29.5\% of the parameters of 3DGS.

\begin{table}[!t]
    \centering
    \caption{Ablations experiments for each component.}
    \vspace{-2mm}
    \renewcommand{\arraystretch}{1.1}
    \setlength{\tabcolsep}{4mm}
    \scalebox{1}{\begin{tabular}{l|ccc}
    \hline\hline
    Method & PSNR & SSIM & LPIPS\\ 
    \hline 
    TG & 26.80 & 0.79 & 0.43\\
    TG + AME & 27.24 & 0.81 & 0.40\\
    TG + AME + Loss & \cellcolor{lightyellow}27.78  & \cellcolor{lightyellow}0.83 & \cellcolor{lightyellow}0.33\\
    TG + AME + Loss + TDS & \cellcolor{lightred}29.36 & \cellcolor{lightred}0.87 & \cellcolor{lightred}0.25\\
    \hline  \hline
    \end{tabular}}
    \vspace{-1mm}\label{tab:ablation}
\end{table}

\label{sec:experiments}


\section{Conclusion}
\label{sec:6_conclusion}

We present \methodName, a compact framework for underwater 3D reconstruction and color restoration. It utilizes Tensorized Gaussians and a physically-based Adaptive Medium Estimation (AME) model to capture object-medium interactions with low-rank factorization for parameter efficiency. Specialized densification strategies and an underwater-specific loss ensure efficient optimization and realistic object recovery. This achieves high-fidelity rendering with fewer parameters, suited for real-time multimedia applications in adverse environments.

\section{Acknowledgments}
This work was supported in part by the National Natural Science Foundation of China under Grant 62461018; in part by the Hainan Provincial Natural Science Foundation of China under Grant No. 625YXQN594; in part by the  Innovation Platform for "New Star of South China Sea" of Hainan Province under Grant No. NHXXRCXM202306; in part by the Specific Research Fund of the Collaborative Innovation Center for Flexible Talents of Hainan Province under Grant No.YSPTZX202410; in part by the Zhongguancun Academy (Grant No. C20250510).

\bibliographystyle{IEEEbib}
\bibliography{main}


\end{document}